\documentclass{article}

\usepackage{wrapfig}
\usepackage{graphicx}
\usepackage{float}          
\usepackage[nonatbib,preprint]{neurips_2023}
\usepackage[utf8]{inputenc} 
\usepackage[T1]{fontenc}    
\usepackage{hyperref}       
\usepackage{url}            
\usepackage{booktabs}       
\usepackage{amsfonts}       
\usepackage{nicefrac}       
\usepackage{microtype}      
\usepackage{xcolor}         
\usepackage{graphicx}

\title{Enhancing ML model accuracy for Digital VLSI circuits using diffusion models: A study on synthetic data generation.}

\author{%
  Prasha Srivastava \\
  IIIT, Hyderabad, India \\
  \texttt{prasha.srivastava@research.iiit.ac.in} \\
  \And
  Pawan Kumar \\
  IIIT, Hyderabad, India \\
  \texttt{pawan.kumar@iiit.ac.in} \\
  \And
  Zia Abbas \\
  IIIT, Hyderabad, India \\
  \texttt{zia.abbas@iiit.ac.in} \\
}

\begin{document}

\maketitle

\begin{abstract}
  Generative AI has seen remarkable growth over the past few years, with diffusion models being state-of-the-art for image generation. This study investigates the use of diffusion models in generating artificial data generation for electronic circuits for enhancing the accuracy of subsequent machine learning models in tasks such as performance assessment, design, and testing when training data is usually known to be very limited. We utilize simulations in the HSPICE design environment with 22nm CMOS technology nodes to obtain representative real training data for our proposed diffusion model. Our results demonstrate the close resemblance of synthetic data using diffusion model to real data. We validate the quality of generated data, and demonstrate that data augmentation certainly effective in predictive analysis of VLSI design for digital circuits. 
\end{abstract}

\section{Introduction}
In recent times, generative AI has experienced remarkable growth, propelled by advances in machine learning (ML) models like Generative Adversarial Networks (GANs) \cite{b2,sachin2023}, Variational Autoencoders (VAEs) \cite{b4}, and the most recent addition,  Denoising Diffusion Probabilistic Models (DDPM) \cite{b3}. These models excel at crafting realistic, high-quality data, gaining traction across domains like image synthesis, text generation, and music composition. This surge in generative AI holds immense potential to reshape ML in diverse fields, including VLSI design, testing, and optimization, particularly in data-sparse scenarios.

Diffusion models have demonstrated excellence in computer vision, natural language processing, and interdisciplinary realms such as medical image reconstruction.   
Precise diffusion models generating high-quality artificial circuit data can revolutionize ML-driven VLSI tasks—performance assessment, design, and testing \cite{b5,b6,b7,b8,b9,b10,b11,b12}. An effective data augmentation approach will help addresses concerns such as privacy, proprietary data access, computation costs, and data acquisition constraints. 

Our study focuses on practical application of VLSI circuit design. We study the effectiveness of synthetic data generation method for VLSI circuit data using Denoising Diffusion Probabilistic Model (DDPM). We show that diffusion models can generate high-quality samples even for circuit data. We present a detailed analysis of their performance over delay estimation of twelve fundamental 22nm CMOS technology-based digital cells, see Table \ref{table:digitaldata}. 

\section{Related Works}
Data scarcity is defined as the inadequacy in quantity or diversity of training data, which can restrict the learning ability of an ML model. It is a universal issue which affects the deployment of AI/ML in multiple domains as discussed by A. Munappy et al. \cite{b13}. Many of the proposed AI/ML applications in VLSI design domain \cite{b8,b10,b12} rely on a large amount of training data. For example in \cite{b12} and \cite{b14} authors have achieved precise training with 15K
and 50K samples respectively. Hence in VLSI domain, training data scarcity stemming from cost, time, and quality constraints poses a challenge which needs to be addressed. Many studies have proposed synthetic data generation to create large scale training datasets in various other domains and provide a significant improvement to existing AI/ML models \cite{b15,b16,b17,b18,b19,b20,b21,b22,b23}. To the best of our knowledge, this is the first time a data augmentation for circuit data is studied with latest diffusion models.

Generative models such as Variational Autoencoders (VAE) \cite{b4}, Generative Adversarial Networks (GAN) \cite{b2}, and Diffusion Probabilistic models \cite{b3} are known to produce large and diverse synthetic data for image datasets. Diffusion models have been observed to have an edge over other generative models in terms of quality for image data generation \cite{b24}. Diffusion models have found applications in various domains ranging from computer vision, natural language generation, temporal data modeling, to interdisciplinary applications in other scientific disciplines \cite{b25}. 

\section{The dataset}
We gathered extensive datasets covering design, process, and performance parameters for 
twelve core digital cells, see Table \ref{table:digitaldata}. These parameters are meticulously chosen to enable subsequent performance assessment using a predictive ML model, validating the utility of dataset.
For training, we employ simulated data from Electronic Design Automation (EDA) tool 
HSPICE \cite{b32}. Training data for Digital cells (Table \ref{table:digitaldata}) comprises vectors of random values, drawn from Gaussian distributions of process parameters. We account for ±10\% variations at 3$\sigma$ in CMOS standard cells at 22nm High-K MGK via Predictive Technology Models (PTM). Twelve process parameters (PMOS and NMOS) are included. In addition to statistical distributions, temperature samples spanning $-55^{\circ}C$  to $125^{\circ}C$  and supply voltage deviations of ±10\% from the nominal (0.8V) are integrated. Load capacitance varies similarly to process parameters. We perform propagation delay estimations via HSPICE Monte-Carlo simulations to obtain training data, encompassing PVT (Process, Voltage, Temperature) variations.

\begin{table}
  \caption{List of digital circuit datasets used in this work.(\textbf{Input parameters for evaluation:} Supply voltage, Temperature, Channel length, Transistor width, Physical and electrical equivalent of oxide thickness, Nominal gate oxide thickness, Source/Drain junction depth, and Channel doping concentration, Load capacitance; \textbf{Output parameters for evaluation:} Propagation Delays)}
  \label{table:digitaldata}
  \centering
  \begin{tabular}{ll|ll}
    \toprule
    Dataset  & Parameters & Dataset  & Parameters \\ 
    \midrule
    NOT gate delay   & 17 & Three input AND-OR circuit delay    & 21\\ 
    Two input NAND gate delay   & 19 & Full adder delay    & 21 \\  
    Two input AND gate delay    & 19 & 2:1 Multiplexer delay    & 21 \\ 
    Two input NOR gate delay    & 19 & Three input NAND gate delay    & 21\\ 
    Two input OR gate delay    & 19 & Three input AND gate delay    & 21  \\ 
    Two input XOR gate delay    & 19 & Three input NOR gate delay    & 21 \\ 
    \bottomrule
  \end{tabular}
\end{table}
\section{Synthetic circuit-data generation using Diffusion Models}
Parametric data from VLSI designs is continuous data that holds valuable insights for ML-driven automation of VLSI performance assessment, design, and testing.
As mentioned before, this paper aims to apply denoising diffusion probabilistic models for generating synthetic data for VLSI designs, enhancing the training of ML models in data-scarce scenarios. This indirectly advances automation in VLSI design tasks. Our work illustrates the development of an accurate diffusion model-based synthetic data generation method for delay estimations in various 22nm CMOS technology-based digital VLSI circuits. The methodology can be divided broadly into the following steps
\subsection{Formulation of a Denoising Diffusion Probabilistic Model tailored for VLSI circuit-data}
\label{formulation}
Diffusion models define a Markov chain of diffusion steps to slowly add random noise to data and then learn to reverse the diffusion process to construct desired data samples from the noise. Diffusion models have two processes to follow:

\textbf{Forward Process} - Here, random noise is incrementally added to data over multiple time steps. This sequential noise introduction is mathematically characterized as follows:
 $x_t = \sqrt{1 - \beta_t} \cdot x_{t-1} + \sqrt{\beta_t} \cdot \epsilon $. 
Where \(x_t\) represents the data at time step \(t\), where \(0<t<T\) and \(T\) are the total number of steps. This process starts with the original data \(x_0\) and iteratively adds noise over \(T\) steps, with \(\beta_t\) controlling the amount of noise added at each step. The variance schedule \(\beta_t\) determines the trade-off between introducing noise and maintaining data fidelity. As \(t\) increases, the noise contribution becomes more significant due to the increasing value of \(\beta_t\).

\textbf{Reverse process} – This phase involves predicting the noise added to each data point during the forward process. A neural network, denoted as \(f_{\theta}\), predicts noise \(\epsilon_t\) from noisy data \(x_t\) as $\epsilon_t = f_{\theta}(x_t)$ 

\textbf{Generating new data} - New data samples are generated by performing the reverse process on random noise samples \(\epsilon_t\) drawn from \(N(0, I)\). The reverse process reconstructs data by iteratively removing predicted noise contributions: $x_{t-1} = \frac{x_t - \sqrt{\beta_t} \cdot \epsilon_t}{\sqrt{1 - \beta_t}} 
$.
The number of steps \(T\) controls the balance between data fidelity and noise injection, influencing the quality of generated samples.

Existing models deal with very complex data modalities such as images. For image generation a UNET architecture with residual connections was proposed. Since our target dataset is relatively less complex than images, we propose a simple encoder-decoder architecture \cite{b28} instead of a UNET for reverse denoising process. 

\subsection{Qualitative evaluation of generated artificial data}
\label{evaluation}
The quality assessment of synthetic data involves evaluating measures like inception score, Frechet inception distance, average log-likelihood, Parzen window estimates, and visual fidelity. However, these metrics primarily cater to image data, making it unclear which measure is optimal for other data modalities. 
Theis et al. \cite{b27} suggested that evaluation for generative models should align with the intended application.

\begin{figure}
  \centering
  \includegraphics[scale=0.32]{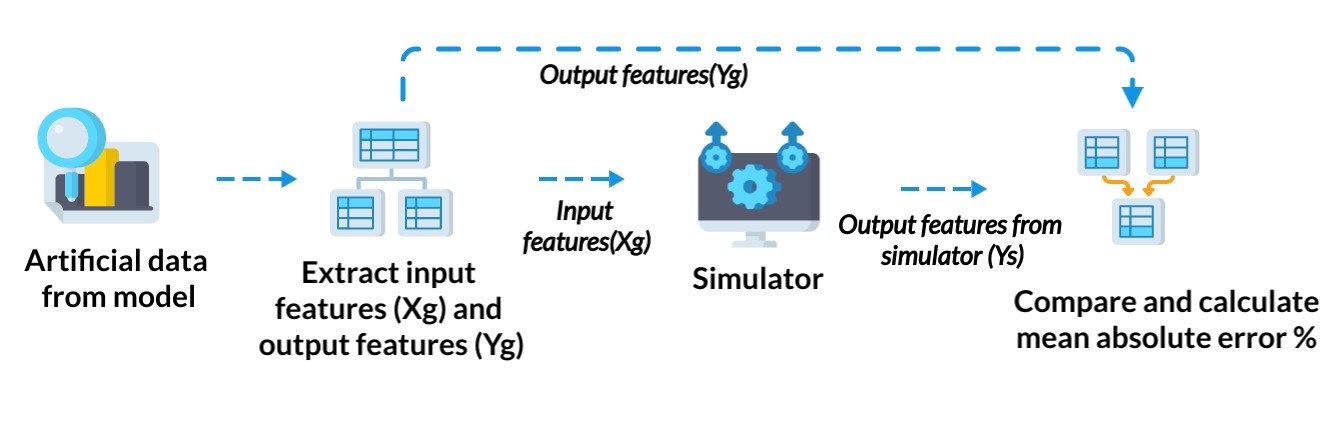}
  \caption{\label{figure:56}Evaluation process for artificially generated data.}
\end{figure}

Thus, we evaluate diffusion models for circuit data by directly comparing them to the source of our data, which is the simulator. We extract specific features from the synthetic dataset, designating them as input features, while the rest are deemed output features. Subsequently, we feed the input features into the simulator and compare the resulting output feature values with the generated output feature values. Our evaluation metric in this study is the mean absolute percentage error. Refer to Figure \ref{figure:56} for a visual representation of the comprehensive evaluation procedure for the generated data.

The diffusion model is trained using just 500 real data samples. Subsequently, artificial samples are generated and used for performance evaluation. Training continues through epochs until desired performance is reached, with hyperparameter adjustment for subpar results. 

\section{Experimental setup and model architecture}
We use {\tt Python-3.8.16} and Google Colab for the training of Diffusion Denoising Probabilistic models. Moreover, our implementation uses {\tt Keras-2.9.0} and {\tt Tensorflow-2.9.2}.
As discussed in \ref{formulation}, a diffusion model is devised for each dataset, encompassing forward and reverse processes for circuit data. For the forward process, we adopt a variance of \(\beta_t\), transitioning linearly from 0.001 to 0.02, following the approach by Ho et al. \cite{b3} The reverse process is realized using an encoder-decoder architecture \cite{b28} with continuous batch normalization. The model utilizes Leaky ReLu activation.
Training data undergoes noise addition through the forward process, gradually transforming to pure noise, and the reverse process learns to de-noise and predict original distribution. After training the diffusion model, the synthetic data is then generated by first sampling a pure random noise then using trained reverse diffusion model to generate the desired sample.

\section{Results}

\begin{figure}
 \begin{center}
  \includegraphics[scale=0.38]{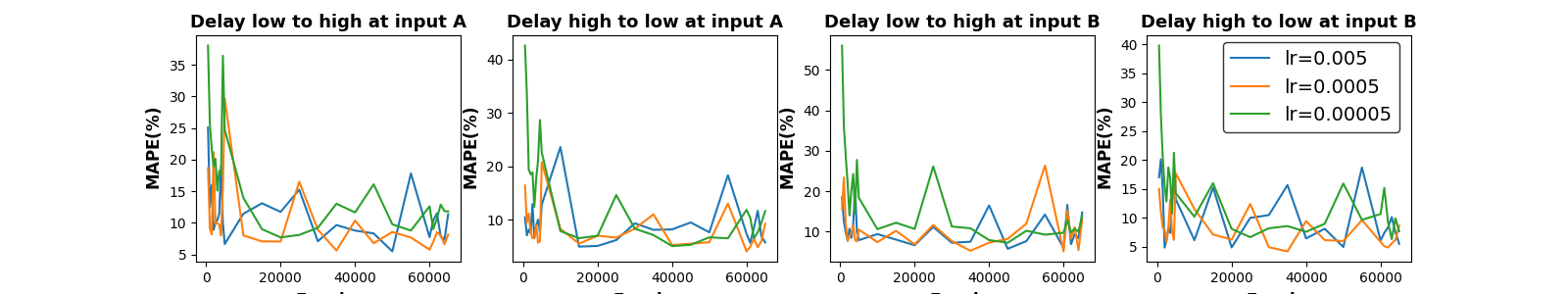}
  \caption{\label{figure:2}Performance of model with different learning rates w.r.t. HSPICE for delay in AND gate dataset. It can be seen that learning rate of 0.0005 ensures low percentage errors across all features.}
  \end{center}
\end{figure}

\begin{figure}
 \centering
  \includegraphics[scale=0.28]{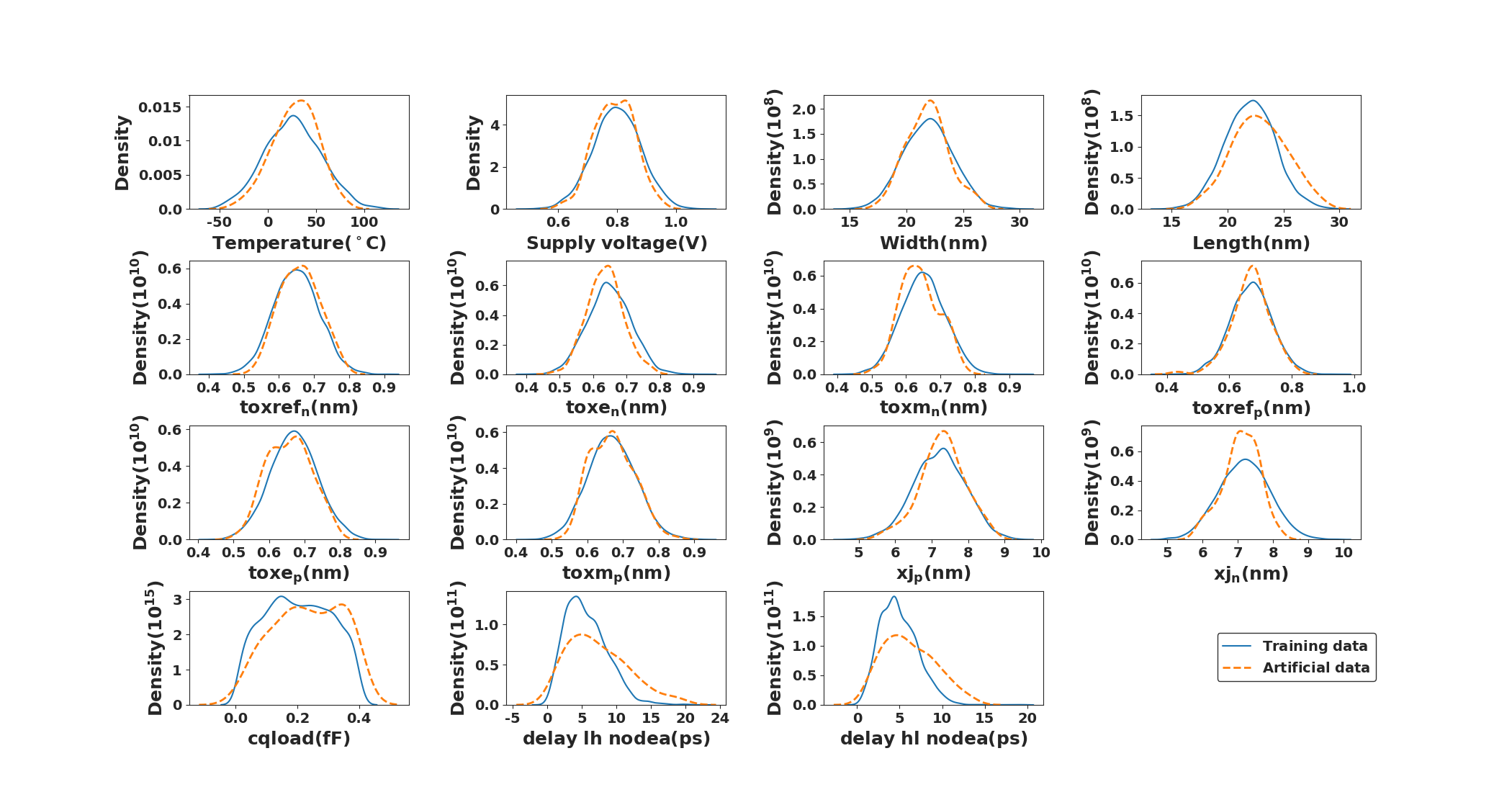}
  \caption{\label{figure:3}Distribution plots of artificially generated data and original data for delay dataset from NOT gate showing close resemblance between two.}
\end{figure}

As discussed in \ref{evaluation}, the output feature values from the model and the simulator are compared to get the right idea of performance. We start the hyper-parameter search by finding the optimal number of layers. Table \ref{table:NOT_layers} depicts the model's performance across varying hidden layer counts. A five-layer architecture emerges as the best choice for the NOT gate dataset featuring 17 attributes. This architecture also holds well for datasets up to 19 attributes.
It was observed that for datasets featuring 21 attributes, a six-layered architecture boosts the learning capacity effectively.
Our subsequent exploration involves different learning rates. Figure \ref{figure:2} indicates that a learning rate near 0.0005 ensures consistently low percentage errors across all features. The Table \ref{table:errordigitaldata} provides the mean absolute percentage errors (MAPE) attained for all datasets, it can be seen that low MAPE is obtained for the various datasets. Additionally, Figure \ref{figure:3} showcases that density distribution of generated and original data for the NOT dataset exhibit a high degree of proximity. Furthermore Table \ref{table:GBR_comparison} shows a significant improvement in a gradient-boosting regression (GBR) model using artificial data, to predict CMOS NOT gate delays. Thus, validating the proposed approach's efficacy in predicting parameters that concern digital circuit design. For brevity, the data generation with other generative models such as GANs or VAEs were not as effective, and hence not shown in results.

\begin{table}
    \caption{Comparison of performance of models with different layers across all features.}
    \label{table:NOT_layers}
    \centering
    \begin{tabular}{lll}
    \toprule
    No. of layers  & Avg. of MAPE(\%)\\ 
    \midrule
    6 hidden layers   & 12.5  \\
    \textbf{5 hidden layers}    & \textbf{3.51}  \\
    4 hidden layers    & 10.5  \\ 
    \bottomrule
    \end{tabular}
\end{table}

\begin{table}
  \caption{Percentage error obtained for different digital circuit datasets used in this work. Here, the error is calculated with respect to the HSPICE. Here A=delay lh node a, B=delay hl node a, C=delay lh node b, D=delay hl node b, E=delay lh node c, F=delay hl node c.}  
  \label{table:errordigitaldata}
  \centering
  \begin{tabular}{llllllll}\toprule
    Delay dataset  & \multicolumn{6}{c}{Mean Absolute Percentage Error}  \\ 
    \cline{2-7}{} & A & B & C & D & E & F \\ 
    \midrule
    NOT gate& 3.9& 3.12& -& - & - & - \\
	Two input NAND gate& 4.41& 6.3& 4.54& 7.52& - & - \\
	Two input AND gate& 5.76& 4.12& 5.13& 5.84& - & - \\
	Two input NOR gate& 5.14& 2.52& 3.92& 6.05&  - & - \\
	Two input OR gate& 3.77& 3.5& 5.57& 4.42&  - & - \\
	Two input XOR gate& 0.34& 3.44& 4.04& 2.94&  - & - \\
	Three input AND-OR circuit& 7.96& 4.83& 4.74& 8.33& 4.3& 8.55\\
	FULL ADDER& 2.85& 2.85& 3.62& 5.93& 2.15& 4.42\\
	2:1 MULTIPLEXER& 3.81& 3.27& 2.91 & 5.53 & 3.68 & 4.03\\
	Three input NAND gate& 7.73 & 4.66 & 6.15 & 5.404& 7.37 & 3.26\\
	Three input AND gate& 4.04 & 4.64 & 5.33 & 2.92 & 3.91 & 3.02\\
	Three input NOR gate& 4.57 & 5.007 & 5.11 & 6.46 & 3.74 & 5.13\\
    \bottomrule
  \end{tabular}
\end{table}

\begin{table}
            \caption{Comparison of performance of a predictive gradient boosting regression model with and without artificial data. (A higher R2 score denoted by $\uparrow$ and a lower MSE, RMSE, MAE and MAPE denoted by $\downarrow$ are preferred.)}
            \label{table:GBR_comparison}
            \centering
            \begin{tabular}{llllllll}
                \toprule
                Metric  & \multicolumn{2}{c}{Real data} & \multicolumn{2}{c}{Real + Artificial data} & \multicolumn{2}{c}{Improvement (\%)}  \\
                \cline{2-7}
                {} & delay lh & delay hl & delay lh & delay hl & delay lh & delay hl  \\ 
                \midrule
                R2 score   & 0.93& 0.95& 0.976& 0.973& 4.95 $\uparrow$& 2.42 $\uparrow$ \\
                MSE    & 4.58$\times10^{-25}$& 2.22$\times10^{-25}$& 2.25$\times10^{-25}$& 1.38$\times10^{-25}$& 50.87$\downarrow$& 37.84$\downarrow$ \\
                RMSE    & 6.77$\times10^{-13}$& 4.71$\times10^{-13}$& 4.75$\times10^{-13}$& 3.72$\times10^{-13}$& 29.84$\downarrow$& 21.02 $\downarrow$\\ 
                MAE    & 5.16$\times10^{-13}$& 3.32$\times10^{-13}$& 3.44$\times10^{-13}$& 2.52$\times10^{-13}$& 33.33$\downarrow$& 24.10$\downarrow$ \\
                MAPE    & 0.106& 0.103& 0.099& 0.083& 6.60$\downarrow$& 19.42$\downarrow$ \\
                \bottomrule
            \end{tabular}
\end{table}

\section{Conclusion}
Achieving model accuracy hinges on high-quality training data, yet obtaining ample data for electronic circuits can be expensive or practically difficult to obtain. 
 
Our study proposes a variant of diffusion model for generating a) synthetic data, b) validation method, and c) subsequently predicting delay estimations in 22nm CMOS technology-based digital VLSI circuits. 

More specifically, the training data is obtained by various simulations in the HSPICE with 22nm CMOS technology nodes.
The technique's efficacy is proven through extensive experiments on twelve essential digital circuit designs, showcasing notably low mean absolute percentage errors with respect to HSPICE circuit simulator. It is also observed that artificial data distribution closely resembles the original data distribution. Improvement in a GBR’s performance using the proposed augmented data is also demonstrated.

\bibliographystyle{ieeetr}
\bibliography{cas_refs}

\end{document}